\def\year{2023}\relax
\documentclass[letterpaper]{article} % DO NOT CHANGE THIS
\usepackage{aaai23}  % DO NOT CHANGE THIS
\usepackage{times}  % DO NOT CHANGE THIS
\usepackage{helvet}  % DO NOT CHANGE THIS
\usepackage{courier}  % DO NOT CHANGE THIS
\usepackage[hyphens]{url}  % DO NOT CHANGE THIS
\usepackage{graphicx} % DO NOT CHANGE THIS
\usepackage{natbib}  % DO NOT CHANGE THIS AND DO NOT ADD ANY OPTIONS TO IT
\usepackage{caption} % DO NOT CHANGE THIS AND DO NOT ADD ANY OPTIONS TO IT
\DeclareCaptionStyle{ruled}{labelfont=normalfont,labelsep=colon,strut=off} % DO NOT CHANGE THIS
\usepackage{xcolor}
\usepackage{tabularx}
\usepackage{amsmath}

\title{Neuro-Symbolic Bi-Directional Translation - Deep Learning Explainability for Climate Tipping Point Research}
\author {
    Chace Ashcraft,\textsuperscript{\rm 1}
    Jennifer Sleeman,\textsuperscript{\rm 1}
    Caroline Tang,\textsuperscript{\rm 1}
    Jay Brett,\textsuperscript{\rm 1}
    Anand Gnanadesikan\textsuperscript{\rm 2}
}
\affiliations {
    \textsuperscript{\rm 1} Johns Hopkins University Applied Physics Laboratory\\
    11100 Johns Hopkins Road Laurel, Maryland 20723\\
    \textsuperscript{\rm 2} Johns Hopkins University\\
    3400 N. Charles St. Baltimore, MD 21218-2683\\
   \{chace.ashcraft,jennifer.sleeman,caroline.tang,jay.brett\}@jhuapl.edu,gnanades@jhu.edu
}

\begin{document}

\maketitle

\begin{abstract}
In recent years, there has been an increase in using deep learning for climate and weather modeling.  Though results have been impressive, explainability and interpretability of deep learning models are still a challenge.  A third wave of Artificial Intelligence (AI), which includes logic and reasoning, has been described as a way to address these issues. Neuro-symbolic AI is a key component of this integration of logic and reasoning with deep learning.  In this work we propose a neuro-symbolic approach called \textit{Neuro-Symbolic Question-Answer Program Translator}, or \textit{NS-QAPT}, to address explainability and interpretability for deep learning climate simulation, applied to climate tipping point discovery. The NS-QAPT method includes a bidirectional encoder-decoder architecture that translates between domain-specific questions and executable programs used to direct the climate simulation, acting as a bridge between climate scientists and deep learning models.  We show early compelling results of this translation method and introduce a domain-specific language and associated executable programs for a commonly known tipping point, the collapse of the Atlantic Meridional Overturning Circulation (AMOC).
\end{abstract}

\section{Introduction}
The abundance of climate-related natural disasters~\cite{botzen2019economic, coronese2019evidence, jafino2020revised}, weather extremes~\cite{extremeweather1, extremeweather2}, and poor air quality~\cite{JACOB200951, nolte2018potential} in recent years has created a sense of urgency in the development of new methodologies for climate research that reduces computation requirements and enables better forecasting, tolerant of a changing environment.  Artificial Intelligence (AI) and in particular, deep learning has shown promise in recent years in both data-driven models~\cite{singh2021deep} and those which incorporate physical and dynamical properties~\cite{fourcastnet}.  However, these methods still tend to suffer from poor interpretability and lack of explainability~\cite{garcez2020neurosymbolic}.  In climate and weather related research both of these properties are critical, as often forecasts influence guidance to the general public and policy makers. 

As described by Garcez et al.\cite{garcez2020neurosymbolic}, the third wave of AI includes deep learning and symbolic representation, described as neuro-symbolic. By incorporating symbolic representation, the black box properties of deep learning models can be informed by a logical understanding of the input and output of the model.  Neuro-symbolic refers to the hybridization of symbolic reasoning or computation techniques with deep learning methods. The strengths of deep learning, such as complex pattern recognition and sequence prediction can be augmented by AI methods such as graph-based search and logic systems to produce systems capable of generating robust, human-interpretable predictions that provide a means of explainability.
 
In this work, we describe a neuro-symbolic model, called \textit{Neuro-Symbolic Question-Answer Program Translator}, or \textit{NS-QAPT}, which symbolically represents deep learning problems and links these representations to natural language. NS-QAPT is a bi-directional translator, that converts natural language questions into surrogate climate model programs and surrogate climate model programs (generated by a deep learning climate simulator) into natural language questions with associated answers.  NS-QAPT was designed to bridge the gap between climate tipping point researchers and deep learning models. 

To test this methodology, we applied the neuro-symbolic translator and deep learning climate simulator to a known climate problem, the collapse of the Atlantic Meridional Overturning Circulation (AMOC).  We demonstrate how NS-QAPT could be integrated with a deep learning climate simulator by using a climate tipping point Generative Adversarial Network (TIP-GAN) \cite{tip-gan} for climate simulation.  Though TIP-GAN could be used to significantly reduce the parameter space by discovering combinations of parameters that lead to models which result in a tipping point, climate researchers need a way to interact with this model and interpret what has been learned.  By combining NS-QAPT with TIP-GAN, climate researchers are able to ask natural language questions of what is learned by TIP-GAN, enabling them to potentially direct their own climate research to smaller parameter spaces.  We evaluated NS-QAPT using a common neuro-symbolic dataset CLEVR \cite{johnson2017clevr} in our previous work~\cite{sleeman_aaai_fall_symposium_2022_paper} and in this work we evaluate NS-QAPT with a custom AMOC-specific question program translation language.

\section{Background--The AMOC}
The AMOC is a globally circulating current in the Atlantic ocean characterized by warm surface water flowing northward, then cooling, sinking, and flowing back southward. The cooling and increase of salinity of ocean water as it flows northward increases the density of surface water, causing it to sink. It then slowly moves southward along the ocean floor until it can rise in the Pacific and Indian oceans. The northern flow of ocean water from the equator is a significant source of heat energy in the northern hemisphere. 

In general, the AMOC plays an important role in the global climate. Small changes to its strength can have potentially global effects, such as significant cooling in the northern hemisphere and changes in precipitation. Some models suggest that the AMOC could weaken or even collapse in the near future~\citep{thornalley2018anomalously,jackson2018hysteresis}, consequences of which may include food insecurity~\cite{benton2020running} and sea level rise~\cite{bakker2022ocean}.

\subsection{AMOC Box Models}

\begin{figure} 
\centering
\includegraphics[width=.95\columnwidth]{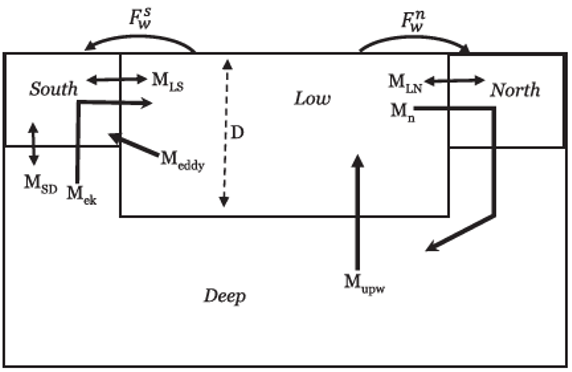}
\caption{The Gnanadesikan Four Box Model of the AMOC}
\label{fig:fourbox}
\end{figure}

Large climate systems are sometimes reduced to surrogate models such as a box model~\citep{levermann2010atlantic}, which simplifies some of the more complex details of the system while maintaining its essential characteristics. This allows the model to theoretically represent the dynamics of their larger counterparts, but are reduced enabling research that would otherwise be computationally infeasible. To experiment with the AMOC and identify states when the AMOC may collapse, we use a four box model from Gnanadesikan et. al.~\cite{gnanadesikan2018flux}, re-implemented in Python\footnote{\url{https://github.com/JHUAPL/PACMANs}}, as a surrogate for a larger global model. A high-level figure of the box model is shown in Figure~\ref{fig:fourbox}. \textit{South} and \textit{North} refer to segments of surface water in those latitudes of the Atlantic ocean. The \textit{Low} box similarly represents the surface water in-between, and the \textit{Deep} box represents all deep water flow. $M_n$ refers to the mass transported through the northern box, and is the primary measure of the AMOC's strength. $F_w^s$ and $F_w^n$ are the freshwater fluxes in the southern and northern boxes, respectfully. Due to warming climate, it is possible for these fluxes to grow due to the melting of ice in each region. The influx of freshwater into the ocean affects the salinity of water, potentially perturbing the the whole system. Freshwater flux perturbations are one possible contributor to eventual AMOC collapse. 

\section{Related Work}

As discussed in \cite{garcez2020neurosymbolic}, neuro-symbolic methods are not necessarily new. Khsola and Dillon published a taxonomy of neuro-symbolic approaches along with a neuro-symbolic system called GENUES, which was designed for real-time alarm processing and based some of their previous work \cite{khosla1993combined, khosla1998welding}. Another example of early neuro-symbolic work is \cite{neagu2002modular}, which attempts to fuse artificial neural networks and fuzzy logic. At that time, neural network models generally consisted of shallow, one or two-layer, perceptron models with less than 100 nodes in the hidden layers and perhaps tens of thousands of parameters~\cite{lawrence1998size, canziani2016, ann_history2016}. AlexNet~\cite{NIPS2012_c399862d} consists of over 60 million parameters. Continued advancements both in neural network architecture design and faster compute have been integral to the success of deep learning. Early attempts at applying to neuro-symbolic techniques to climate include forecasting red tides~\cite{fdez2003forecasting} and energy management~\cite{velik2011towards, velik2013}. Fdez-Riverola uses neural networks to index into a case-based reasoner, which stores latent representations of previous fuzzy logic rules that predicted accurate phytoplankton concentrations in the past. Velik uses neural networks to help covert raw sensor information into aggregate latent information, which is then used by a rule-based planning and control module to regulate the power consumption of the electrical devices in a home.

While recent applications of deep learning to climate are becoming plentiful~\cite{rasp2018deep, reichstein2019deep, bury2021deep, singh2021deep, schultz2021can}, and climate related neuro-symbolic work with shallow neural networks exists, we were not able to find recent applications of neuro-symbolic (i.e. that use deep learning) methods applied to climate change--and, more specifically, to climate tipping points--in our literature survey.

\section{Model Design}
NS-QAPT is inspired by the neuro-symbolic Concept Learner (NS-CL) by Mao et. al.~\cite{nscl} and CLEVRER~\cite{yi2019clevrer}. The NS-CL learns to associate latent representations of objects in a scene with given concept words from a domain specific language (DSL), as well as learning to manipulate and execute quasi-symbolic programs to answer questions about the scene. The authors leverage the CLEVR dataset~\cite{johnson2017clevr}, which consists of sets of computer generated images of objects, questions about the relationships between objects in the images, and corresponding ``programs'' that answer the given questions. First, the NS-CL uses a perception network to extract latent representations of objects in an image. Then the extracted representations are given to a reasoning module, which identifies the concepts represented in the latent representations and what operations to perform on each concept to answer the question. The reasoning operations consist of implementations of the programs given in the CLEVR dataset. Training is done end-to-end using stochastic gradient decent and the REINFORCE~\cite{williams1992simple} algorithm. 

NS-QAPT differs from NS-CL \cite{nscl} or CLEVRER \cite{yi2019clevrer} in that it does not require a perception module to extract concepts from an image since this is a text-only problem. NS-QAPT is also a bidirectional translation between natural language and programs, which is not part of NS-CL \cite{nscl} or CLEVRER \cite{yi2019clevrer} methodology. NS-QAPT learns programs in a purely sequence-to-sequence manner, rather than from search or reinforcement learning.

% \begin{figure}
%     \centering
%     \includegraphics[width=0.95\columnwidth]{images/neuro.png}
%     \captionsetup{width=.98\columnwidth}
%     \caption{NS-QAPT's bidirectional text-to-program translation architecture. NS-QAPT is a combination of an auto-encoder and two encoder-decoder models. All encoders and decoders share a latent space.}
%     \label{fig:neuro}
% \end{figure}

\begin{figure*}
    \centering
    \includegraphics[width=0.99\textwidth]{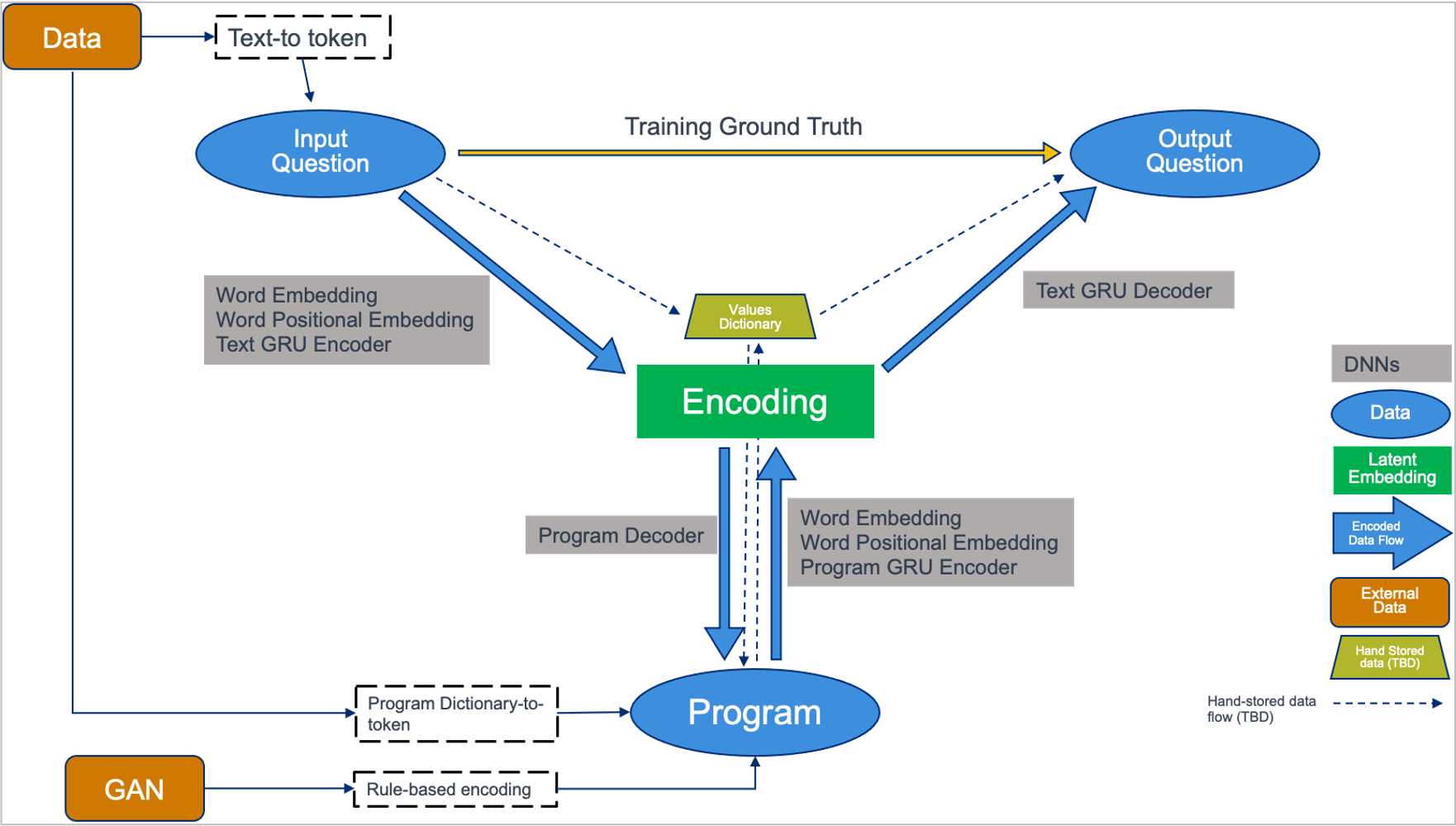}
    \captionsetup{width=.98\textwidth}
    \caption{NS-QAPT's bidirectional text-to-program translation architecture. NS-QAPT is a combination of an auto-encoder and two encoder-decoder models. All encoders and decoders share a latent space.}
    \label{fig:neuro}
\end{figure*}

The bidirectional question-to-program translation is accomplished via a triangular shaped system of model architectures as seen in Figure~\ref{fig:neuro}. The three pieces of NS-QAPT include a question-to-question (QTQ) auto-encoder, a question-to-program (QTP) encoder-decoder, and a program-to-question (PTQ) encoder-decoder. All parts share the same latent space and token embedding, and are optimized jointly during training. 

Let $B$ denote a batch of $N$ examples $\mathbf{x}_i = (x_i^Q, x_i^P)$, where $x_i^Q$ is the vector of integer tokens representing the $i^{th}$ natural language question in the batch, and $x_i^P$ is the vector representing the corresponding tokenized program. We denote NS-QAPT's predicted output of $\mathbf{x}_i$ as $\mathbf{\hat{y}}_i = (\hat{y}_i^{QTQ}, \hat{y}_i^{QTP}, \hat{y}_i^{PTQ})$, where $\hat{y}_i^{QTQ}$ is NS-QAPT's question-to-question auto-encoder prediction, $\hat{y}_i^{QTP}$ its question-to-program encoder-decoder prediction, and $\hat{y}_i^{PTQ}$ its program-to-question encoder-decoder prediction. Then ground truth may be written as

$$
\mathbf{y}_i = (y_i^{QTQ}, y_i^{QTP}, y_i^{PTQ}) = (x_i^Q, x_i^P, x_i^Q).
$$

\noindent
Let $L_{CE}$ denote the standard cross-entropy loss, we define the \textbf{total cross-entropy}, $L_{TCE}$, as

% \begin{flalign*}
%     L_{TCE}(\mathbf{\hat{y}}, \mathbf{y}) = &L_{CE}(\hat{y}^{QTQ}, y^{QTQ})&&\\\nonumber 
%         &+ L_{CE}(\hat{y}^{QTP}, y^{QTP})&&\\\nonumber
%         &+ L_{CE}(\hat{y}^{PTQ}, y^{PTQ}).&& 
% \end{flalign*}

\begin{align*}
    L_{TCE}(\mathbf{\hat{y}}, \mathbf{y}) = &L_{CE}(\hat{y}^{QTQ}, y^{QTQ})&&\\\nonumber 
        &+ L_{CE}(\hat{y}^{QTP}, y^{QTP})&&\\\nonumber
        &+ L_{CE}(\hat{y}^{PTQ}, y^{PTQ}).&& 
\end{align*}

\noindent
Let $|v|$ be the length of a vector $v$, and $\Big | \normalsize z \Big |$, be the absolute value of a scalar $z$. We define the \textbf{total length difference}, $L_{TLD}$, as

\begin{align*}
    L_{TLD}(\mathbf{\hat{y}}, \mathbf{y}) = &\Big | \normalsize|\hat{y}^{QTQ}| - |y^{QTQ}| \Big |\normalsize&&\\\nonumber 
    &+ \Big | \normalsize|\hat{y}^{QTP}| - |y^{QTP}| \Big |&&\\\nonumber
    &+ \Big | \normalsize|\hat{y}^{PTQ}| - |y^{PTQ}| \Big |.&&
\end{align*}

\noindent
Finally, let $\alpha$ be a constant scalar. Using this notation, we may write the loss function, $\mathcal{L}$, for NS-QAPT as follows:

\begin{align*}
\mathcal{L}(B) = \frac{1}{N} \sum_i^N L_{TCE}(\mathbf{\hat{y}}_i, \mathbf{y}_i) - \alpha L_{TLD}(\mathbf{\hat{y}}_i, \mathbf{y}_i)&&
\end{align*}

\noindent
where $\alpha=0.001$. In the cases where $L_{TLD}(\mathbf{\hat{y}}, \mathbf{y}) > 0$, the inputs to $L_{TCE}$ are truncated to the length of the shorter of the predicted and ground truth vectors. 

At a high level, the loss on a batch consists of summing the cross-entropy between model predictions and ground truth for QTQ, QTP, and PTQ data, subtracting the absolute values of the differences in sequence lengths between predictions and ground truth, scaled down by a constant factor, and then returning the mean over all $N$ examples in the batch.   

% I just realized that I may not need to differentiate by QTQ, QTP, and PTQ, since \hat{x} for each x will correspond to the correct ground truth. ... In the code each CE Loss is computed per category and then summed, as explained, so I think it does matter in the end? I don't think L_CE(x \cup y, x' \cup y') = L_CE(x, x') + L_CE(y, y').

NS-QAPT's question and program encoders share a sequence representation consisting of a word embedding of size 512 and a modified learned positional embedding~\cite{nscl} of size 128. Both encoders are bidirectional, 2-layer Gated Recurrent Units (GRUs)~\cite{cho2014properties} with a hidden size of 512. The decoders have a similar architectures, each being a single-direction, single-layer GRU with hidden size 1024 followed by two linear layers. The first layer is size 512 and followed by a LeakyReLU~\cite{maas2013rectifier}, and the second is size 253, which is the size of the vocabulary. The vocabulary was constructed from tokens in the CLEVR dataset and tokens from the AMOC questions (For the CLEVR experiment, the vocabulary only consisted of CLEVR tokens, and thus these layers were smaller for our CLEVR experiment.). Currently, numerical values are converted to ``\texttt{VALUE}'' tokens pre-encoding and are stored in a dictionary to be passed to the decoders. \texttt{VALUE} tokens are replaced with numerical values post-decoding in the same order they were encoded. While dealing with numerical values this way works reasonably well, we hope future work will use more advanced methods, such as a learned association with numerical values and their position in the sequence. 

\section{AMOC Dataset}
We evalauted our method using two question-answer-program translation datasets. The first was based on the CLEVR~\cite{johnson2017clevr} dataset, which a well-developed dataset about relationships of geometric objects in images used to benchmark neuro-symbolic methods. We described the results of evaluating our methodology using the CLEVR dataset in our previous work \cite{sleeman_aaai_fall_symposium_2022_paper}.  The second is based on AMOC-related questions and answers, and program translations, created specifically for this work.  

We generated a custom set of AMOC-collapse questions and their corresponding programs to further evaluate our model. Our approach is to define question ``forms'' in which words and numerical values may be inserted to create valid questions answerable by a set of implemented programs. For example, one question form is ``What is the value of M\_n at time step \{1\} if \{2\} is \{3\}?'' where \textit{M\_n} represents the mass transported through the northern box ($M_n$), and \{1\}, \{2\}, and \{3\} are placeholders for possible values. Replacing \{1\} with $4000$, \{2\} with $Fwn$, and \{3\} with $5000$ results in the question:

\begin{quote}
    ``What is the value of M\_n at time step 4000 if $Fwn$ is 5000?''
\end{quote}

\noindent
The corresponding program is:

\vspace{0.1in}

{\raggedright \quad \quad ``FinalValue( \\ 
        \quad \quad \quad four\_box\_model( \\ 
        \quad \quad \quad \quad SetTo(N,4000),SetTo(Fwn,5000)), \\ 
\quad \quad M\_n)''}
\vspace{0.1in}

\noindent
where ``\texttt{four\_box\_model}'' runs the Four Box Model simulation for a vector of parameters, ``\texttt{SetTo}($x$, $y$)'' sets the Four Box Model parameter $x$ to $y$, and ``\texttt{FinalValue}($V$, $z$)'' extracts the data representing the variable $z$ from $V$, the set of outputs of the Four Box Model simulation, and returns value of that data at the final step of the simulation. 

Each question has a discrete set of parameters from which question, program pairs may be generated. If a question requires a numerical entry, a value is generated by adding noise to the default box-model value for the associated parameter, or from a standard normal distribution if the entry does not have a related default box-model value. Noise is also constrained to ensure each value is within a reasonable range for its a parameter. 

Some questions allow for repeated phrases using different words or numbers, creating a combinatorial expansion of similar questions. For example, ``If I set Fwn to 5.8e4, M\_ek to 2.6e7, will M\_n increase?'' could be extended to ``If I set Fwn to 5.8e4, M\_ek to 2.6e7, and D\_low0 to 439, will M\_n increase?'' essentially adding another \texttt{SetTo} call to the \texttt{four\_box\_model} function. Each \texttt{SetTo} call (or clause) can be given in any order, and for as many parameters as desired, creating the combinatoral expansion of questions. For our dataset, we use no more than three parameters per question.

Finally, some phrases may also be substituted with different, synonymous phrases, in order to build more diversity into dataset. For example ``If I set $Fwn$ to'' may be replaced with ``Setting the freshwater flux in the northern ocean to.'' without changing the meaning of the question and thus not changing the corresponding program. We include several such replacements to generate our current dataset.

Table~\ref{tab:questions} shows some example questions from the dataset, with Table~\ref{tab:programs} showing their corresponding programs.

\begin{table}[h]
    \centering
    \small
    \begin{tabular}{|p{0.6\columnwidth}|}
    \hline
        Example  \\ \hline
        What is the value of M\_n at time step 4000 if Fwn is 5000? \\ \hline
        If Fwn is 45113 and M\_ek is 2.7e7, does the AMOC collapse?  \\ \hline
        What is the final value of the AMOC when Fwn is 49243? \\ \hline
        Does Fwn collapse the AMOC at 49483?  \\ \hline
        If I set Fwn to 5.8e4, M\_ek to 2.6e7, and D\_low0 to 439, will M\_n increase? \\ \hline
        If I increase Fwn by 2052, will M\_n increase? \\ \hline
        If I increase Fwn by 720, will salinity in the northern box increase?  \\ \hline
    \end{tabular}
    \captionsetup{width=.9\columnwidth}
    \caption{Examples of question types generated for our AMOC-collapse dataset. Variables from the Four Box model~\cite{gnanadesikan2018fourbox}: \textit{M\_n} (mass transport through the northern box), \textit{Fwn} (freshwater flux in the northern box), \textit{M\_ek} (Ekman transport), and \textit{D\_low0} (start depth of the low box).}
    \label{tab:questions}
\end{table}

\begin{table}[h]
    \centering
    \small
    \begin{tabularx}{0.9\columnwidth}{|p{0.846\columnwidth}|}
    \hline
        Example \\ \hline
        {\raggedright FinalValue( \\ 
        \quad four\_box\_model( \\ 
        \quad \quad SetTo(N,4000),SetTo(Fwn,5000)), \\ 
        M\_n)} \\ \hline
        {\raggedright ChangeSign( \\ 
            \quad four\_box\_model( \\ 
            \quad \quad SetTo(Fwn,45113),SetTo(M\_ek,2.7e7)), \\ 
            M\_n)} \\ \hline
        {\raggedright FinalValue( \\ 
        \quad four\_box\_model(SetTo(Fwn,49243)),M\_n)} \\ \hline
        {\raggedright ChangeSign( \\ \quad four\_box\_model(SetTo(Fwn,49483)),M\_n)} \\ \hline
        {\raggedright IncreaseOf( \\ 
            \quad four\_box\_model( \\
                \quad \quad SetTo(Fwn, 5.8e4), \\ 
                \quad \quad SetTo(M\_{ek}, 2.6e7), \\ 
                \quad \quad SetTo(D\_low0, 439)), 
        \\ M\_n)} \\ \hline
        {\raggedright IncreaseOf( \\ \quad four\_box\_model(IncreaseBy(Fwn,2052)),M\_n)} \\ \hline
        {\raggedright IncreaseOf(\\ \quad four\_box\_model(IncreaseBy(Fwn,720)),S\_north)} \\ \hline
    \end{tabularx}
    \captionsetup{width=.97\columnwidth}
    \caption{Examples of programs generated for our AMOC-collapse dataset. The \textit{four\_box\_model} function represents a run of the Four Box model~\cite{gnanadesikan2018fourbox}. Arguments are model parameters to update from the defaults. We assume the output of this model is AMOC time-series data, including north box mass transport \textit{M\_n}. \textit{S\_north} represents the salinity of the water in the northern box (see Table~\ref{tab:questions} for information on other Four Box model variables).}
    \label{tab:programs}
\end{table}

\section{Experimental Setup and Results}
We describe the experimental setup when applying this methodology to the AMOC question dataset. The metric we report is the normalized Levenshtein distance~\cite{yujian2007normalized}. Levenshtein distance is a distance metric for the number of replacements required to change one sequence into another. Normalized Levenshtein distance converts this measure to be in the interval $[0, 100]$, and like accuracy, greater values mean more similar sequences. Therefore approaching 100 is the desired behavior. 

%, and ``equality'' scores, which refer to token-to-token accuracy between the predicted sequence and ground truth

\subsection{AMOC}

\begin{figure*}
    \centering
    \includegraphics[height=0.3\textheight]{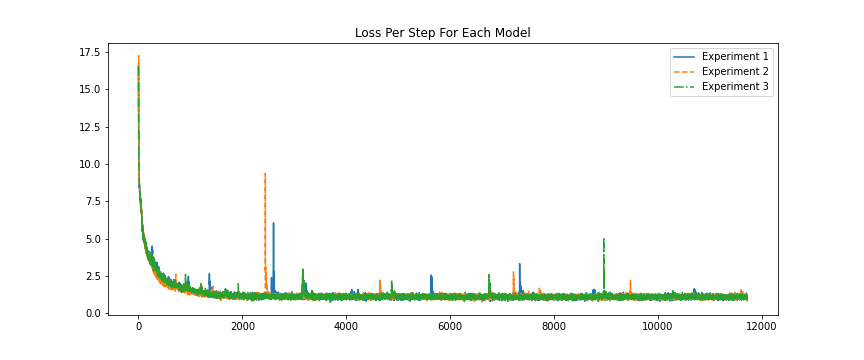}
    \caption{AMOC mean loss per optimization step between three experiments training the translation model.}
    \label{fig:meanallloss}
\end{figure*}

We train NS-QAPT for three epochs on a dataset consisting of 250,000 examples, balanced to have approximately equal numbers of each question and equal numbers of question sequence lengths. We then test on a holdout set of 25,000 examples, which are separated from the training data prior to balancing, and does not contain repeat examples like in the training data in order to maintain balance. We perform this experiment three times with different seeds, and refer to each simply as Experiment 1, Experiment 2, and Experiment 3. As seen in Figure~\ref{fig:meanallloss}, all models converge within the first epoch ($\sim$4000 steps).

In Table~\ref{tab:model_eval_comp}, we also see that the performance between the models is also very similar.

% \begin{table}
%     \centering
%     \begin{tabular}{|l|c|c|}
%         \hline
%         Measure & Mean & StDev \\ \hline
%         Equality Score & 76.76 & 0.77 \\
%         Norm. Levenshtein Distance & 87.51 & 0.912 \\ \hline
%     \end{tabular}
%     \caption{AMOC mean evaluation scores between the three NS-QAPT models. Equality accuracy is accuracy token for token in a sequence. Levenshtein distance is the normalized Levenshtein distance. While a very high-level evaluation, it shows a low variance between training runs.}
%     \label{tab:model_eval_comp}
% \end{table}

\begin{table}
    \centering
    \begin{tabular}{|c|c|c|c|}
        \hline
         % & \multicolumn{3}{c|}{NLD} \\ \hline
        Model & QTQ & QTP & PTQ \\ \hline
        1 & 99.99 & 99.99 & 61.69 \\
        2 & 99.99 & 99.99 & 61.16 \\
        3 & 99.99 & 99.99 & 63.04 \\ \hline
        Mean & 99.99 & 99.99 & 61.97 \\
        StDev & 0.0006 & 0.0002 & 0.79 \\ \hline
    \end{tabular}
    \caption{AMOC mean evaluation scores between the three NS-QAPT models. Normalized Levenshtein distance~\cite{yujian2007normalized} is a normalized measure of the number of replacements required to convert the predicted sequence to the ground truth sequence. NS-QAPT shows low variance between training runs.}
    \label{tab:model_eval_comp}
\end{table}

% \begin{figure}
%     \centering
%     \includegraphics[width=\columnwidth]{model_comps.png}
%     \caption{Mean of mean evaluation scores over three different training instances. Equality accuracy is accuracy token for token in a sequence. Levenshtein distance is the normalized Levenshtein distance. While a very high level evaluation, it shows a low variance between training runs.}
%     \label{fig:model_eval_comp}
% \end{figure}

\subsection{Analysis}

% \begin{table}
%     \centering
%     \begin{tabular}{|l|c|c|c|c|}
%         \hline
%          & QTQ & QTP & PTQ & Overall \\ \hline
%         Dist & 99.998 & 99.989 & 61.693 & 87.227 \\ \hline
%     \end{tabular}
%     \caption{Mean normalized Levenshtein distance for one of the trained NS-QAPT models over all test samples.}
%     \label{tab:overall_bar}
% \end{table}

\begin{figure}
    \centering
    \includegraphics[width=0.98\columnwidth]{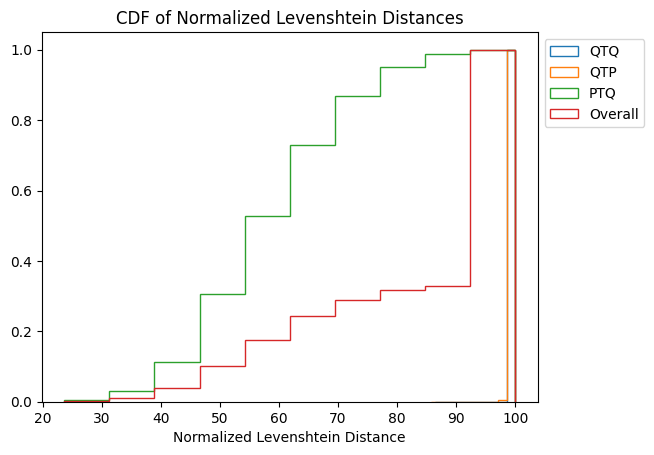}
    \caption{CDF plot of normalized Levenshtein distance performance on AMOC questions for Experiment 1 NS-QAPT models.}
    \label{fig:overall_cdf}
\end{figure}

Figure~\ref{fig:overall_cdf} shows the distribution of normalized Levenshtein distances over all examples in the test dataset. We can see that the QTQ and QTP CDFs are completely concentrated over the 95-100 distances, while PTQ is more spread out over distances from 20 to 100. The overall distribution helps show the contribution of PTQ performance to the overall normalized Levenshtein distance. 

In general, we see that the model performs quite well on the QTQ and QTP portions of evaluation, but struggles with PTQ. For QTQ and QTP, the model appears to reproduce ground truth with almost 100\% accuracy for all the different types of questions, but the normalized Levenshtein distance for PTQ is closer to 70 (100 is perfect). Closer examination shows that this may, in part, be a consequence of the lack of variety in the set of training programs. 23,502 of the 25,000 test questions all came from the same question form. Indeed, as seen in Table~\ref{tab:test_counts}, the consequence of the data generation and train and test split is a very small number of test examples for most questions. 

\begin{figure}
    \centering
    \includegraphics[width=\columnwidth]{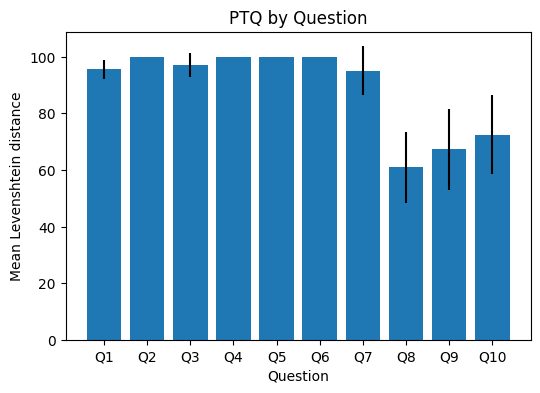}
    \caption{Mean and standard deviation normalized Levenshtein distance by question for Experiment 1 PTQ performance.}
    \label{fig:ptq_by_q}
\end{figure}

Figure~\ref{fig:ptq_by_q}, shows test performance by question for Experiment 1. The NS-QAPT model performs well on questions 1-7, nearing 100 normalized Levenshtein distance, but drops to between 60\% and 80\% for questions 8-10. As these questions constitute more than 99\% of the dataset, the PTQ performance in Table~\ref{tab:model_eval_comp} is close to the performance on just question 8. Taking an unweighted mean of the performance over questions gives a mean normalized Levenshtein distance of \textbf{88.8}, which is significantly higher. 

Questions 8-10 have so many more examples because they their forms allow for multiple calls to the \texttt{SetTo} method, and have several synonymous phrases that many interchanged to produce variants of the same question. However, while this is helpful for producing many questions with subtle variations, the many of these variations produce the same program. The lack of variety in the program set may be the cause of poorer performance on these questions. 

\begin{table}[]
    \centering
    \begin{tabular}{|c|c|c|c|c|}
        \hline
        Q1 & Q2 & Q3 & Q4 & Q5 \\ \hline 
        4 & 26 & 3 & 4 & 67  \\ \hline \hline
        Q6 & Q7 & Q8 & Q9 & Q10 \\ \hline
        4 & 4 & 23,502 & 198 & 1,188 \\ \hline
    \end{tabular}
    \caption{Resulting question counts for the test set after 90/10 train-test split.}
    \label{tab:test_counts}
\end{table}

% A large percentage of questions in the dataset come from one question form.  We did mitigate this bias by balancing the data; however, the programs generated for many of the different questions variants are the same, and programs for different questions are very similar. 

% \begin{figure}
%     \centering
%     \includegraphics[width=0.9\columnwidth]{norm_lev_bar.png}
%     \caption{Mean-of-mean normalized Levenshtein distance over all unique question token lengths.}
%     \label{fig:overall_bar}
% \end{figure}

It is also worth noting that the normalized Levenshtein distance measure does not capture correctness of meaning between predictions and ground truth. For example, consider the following prediction vs versus ground truth for both CLEVR and AMOC PTQ outputs:

\pagebreak

\begin{quote}
\small
Prediction: ``\textbf{if i increase epsilon by 4.24e-06, will temperature in the low latitude box increase?}''

Ground Truth: ``\textbf{by increasing epsilon by 4.24e-06, will temperature in the low latitude box increase?}''

Levenshtein distance \textbf{93.4}
\end{quote}

\noindent Examining various model predictions shows that common mistakes include missing and repeated tokens, in addition to semantically related errors such as synonym substitutions.

\section{Conclusions and Future Work}
Neuro-symbolic methods have the potential to overcome reluctance in using deep learning models for weather and climate forecasting, as they provide a means to interrogate what is learned by the neural methods and a natural language for easy adoption. By coupling the neuro-symbolic method with a deep learning simulator, these methods can work together to reduce the search spaces that are required for climate modeling problems such as AMOC collapse or other types of climate tipping points, potentially enabling faster and more accurate forecasting, with results that are interpretable and explainable. 

We described a neuro-symbolic bi-directional translation model to translate between questions and programs that pertain to a neural simulation built to identify areas in state space that warrant climate modeling exploration as it relates to AMOC collapse. We introduced an AMOC question dataset, and showed how our model is able to translate from questions to programs with a high degree of accuracy and translate from programs to questions with slightly lower accuracy. As we advance the AMOC language further we expect to enable a richer set of questions and improve the program-to-question performance. Future work will also include exploring semantic methods to support one-to-many translations from programs to question. 

\bibliography{cits}

\section{Acknowledgments}
Approved for public release; distribution is unlimited. This material is based upon work supported by the Defense Advanced Research Projects Agency (DARPA) under Agreement No. HR00112290032.

\end{document}